\documentclass[10pt,twocolumn,letterpaper]{article}

\usepackage[pagenumbers]{cvpr} % To force page numbers, e.g. for an arXiv version

\usepackage{graphicx}
\usepackage{amsmath}
\usepackage{amsfonts}
\usepackage{booktabs}

\makeatletter
\renewcommand{\paragraph}{%
  \@startsection{paragraph}{4}%
  {\z@}{0.75ex \@plus 1ex \@minus .2ex}{-1em}%
  {\normalfont\normalsize\bfseries}%
}
\makeatother

\usepackage{mathtools}
\usepackage{cuted} 
\usepackage{capt-of}
\usepackage[font={small}]{caption}
\usepackage{epsfig}
\usepackage{graphicx}
\usepackage{booktabs} % To thicken table lines
\usepackage{multirow}
\usepackage{subcaption}
\usepackage[accsupp]{axessibility}

\usepackage[pagebackref,breaklinks,colorlinks]{hyperref}

\usepackage[capitalize]{cleveref}
\crefname{section}{Sec.}{Secs.}
\Crefname{section}{Section}{Sections}
\Crefname{table}{Table}{Tables}
\crefname{table}{Tab.}{Tabs.}

\def\cvprPaperID{}
\def\confName{CVPR}
\def\confYear{2022}

\begin{document}

\title{Learning to generate line drawings that convey geometry and semantics}

\author{Caroline Chan \qquad
Fr\'edo Durand \qquad
Phillip Isola \\
{\tt\small \{cmchan, fredo, phillipi\}@mit.edu}\\
MIT
}

\maketitle

\begin{strip}\centering
\vspace{-0.7in}
\includegraphics[width=\textwidth]{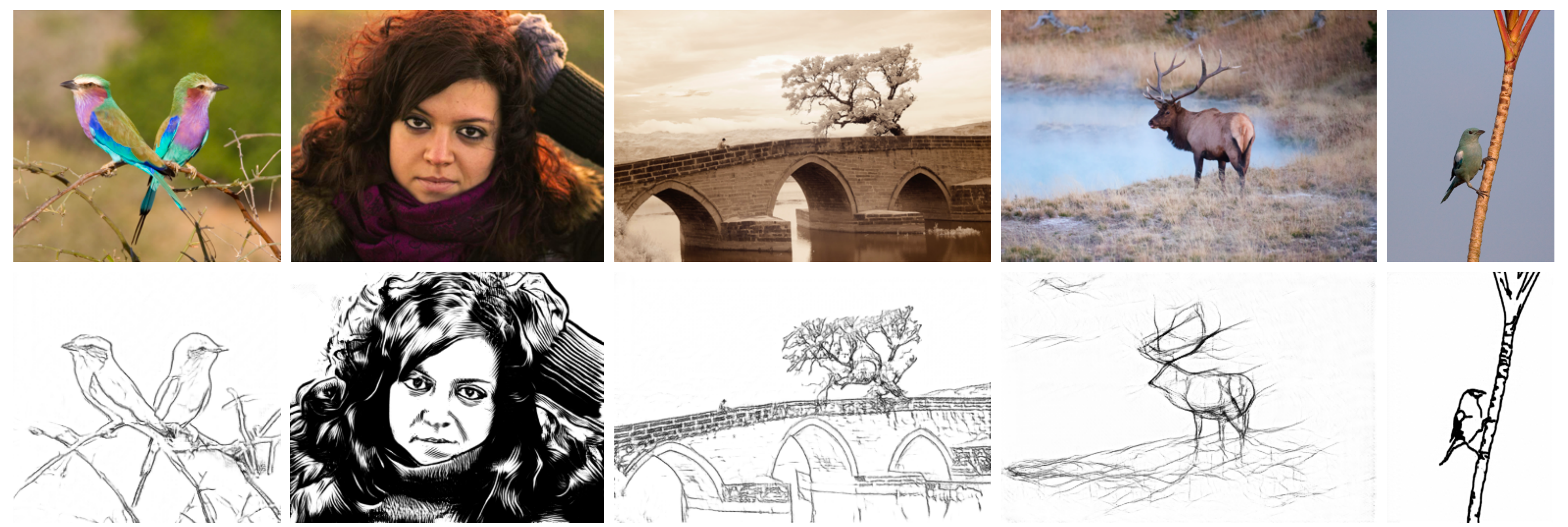}
\vspace{-.2in}
\captionof{figure}{Given a set of photographs, our method is capable of making line drawings in different styles seen above. Our method only requires unpaired data during training.}
\label{fig:feature-graphic} \vspace{-0.1cm}
\end{strip}
\vspace{-0.9cm}

%%%%%%%%% ABSTRACT
\begin{abstract}
\vspace{-0.4cm}
This paper presents an unpaired method for creating line drawings from photographs. Current methods often rely on high quality paired datasets to generate line drawings. However, these datasets often have limitations due to the subjects of the drawings belonging to a specific domain, or in the amount of data collected. Although recent work in unsupervised image-to-image translation has shown much progress, the latest methods still struggle to generate compelling line drawings. We observe that line drawings are encodings of scene information and seek to convey 3D shape and semantic meaning. We build these observations into a set of objectives and train an image translation to map photographs into line drawings. We introduce a geometry loss which predicts depth information from the image features of a line drawing, and a semantic loss which matches the CLIP features of a line drawing with its corresponding photograph. Our approach outperforms state-of-the-art unpaired image translation and line drawing generation methods on creating line drawings from arbitrary photographs. For code and demo visit our webpage \small{\url{carolineec.github.io/informative_drawings}}
\end{abstract}

%%%%%%%%% BODY TEXT
\vspace{-0.6cm}
\section{Introduction}
Through introspection and experimentation, human artists have learned to create line drawings that provide compelling depictions of shape and meaning.  
A longstanding goal of non-photorealistic rendering is to reproduce this feat and, given an input image, to automatically generate line drawings that are effective at conveying geometry and identity. Manually instilling these qualities into computer-generated line drawings is difficult however because the goals are defined in elusive terms of human perception and cognition. Generating line drawings from photographs presents additional challenges: most photographs lack ground-truth geometry data, and often portray complex scenes with multiple subjects and interactions. Naturally, it would make sense to learn from drawings created by humans or to use humans to evaluate automatic line drawing methods. Unfortunately, the creation of such datasets is challenging and scalability is low.

In this paper, we seek to automatically generate effective line drawings from photographs without requiring paired training data and without requiring human judgment of the implied shape. Our key idea is to view the problem as an encoding through a line drawing and to maximize the quality of this encoding through explicit geometry, semantic, and appearance decoding objectives. Our method approaches line drawing generation as an unsupervised image translation problem which uses various losses to assess the information communicated in a line drawing. This evaluation is performed by deep learning methods which decode depth, semantics, and appearance from line drawings. The aim is for the extracted depth and semantic information to match the scene geometry and semantics of the input photographs. Appearance preservation follows from cycle consistency~\cite{zhu2017unpaired,kim2017learning,yi2017dualgan}. With these objectives, our method is able to create convincing line drawings given unpaired data.

Our main contributions are as follows. We present an unsupervised method for automatic line generation which explicitly instills geometry and semantic information into drawings. We apply our method on many styles of line drawings and present results in Section~\ref{reesults}. We also provide analysis of the geometry and semantic information conveyed by our drawings, visual comparisons against several baselines, and an ablation study. 

\section{Related Work}
Line drawings are of particular interest in both art history and psychology. Although studies suggest that the human visual system understands line drawings comparably to photographs~\cite{biederman1988surface,hochberg1962pictorial,kennedy1975outline,hubel1959receptive,hubel1968receptive,yonas1994infants}, it is still unclear why line drawings are effective representations. Several theories exist for this topic, but this area requires further study~\cite{sayim2011line,hertzmann2020line,hertzmann2021role}. 

There has been extensive work on creating line drawings from 3D geometry. Approaches range from applying image processing to depth and normal maps~\cite{canny1986computational,saito1990comprehensible}, using geometric features on top of occluding contours~\cite{benard2019line,ohtake2004ridge,decarlo2003suggestive,judd2007apparent}, to ensembling all geometry-based approaches with deep learning~\cite{liu2020neural}. Although these methods successfully generate line drawings from 3D models, they cannot be applied to arbitrary photographs with unavailable 3D geometry. Furthermore, most methods draw lines in only one style, although Neural Strokes~\cite{liu2021strokes} addresses this issue. Instead, our method creates stylized line drawings from 2D photographs which convey 3D geometry.

Most 2D-based line drawing generation methods rely on supervised data. This includes using ground truth stroke or vector graphics data to create drawings~\cite{ha2018neural,das2020beziersketch,spiral,smirnov2020dps}. This stroke-based approach is often supported by differentiable architectures which can draw lines~\cite{frans2018unsupervised,wang2020sketchembednet,bessmeltsev2019vectorization,huang2019learning,mo2021general,Li:2020:DVG,zheng2018strokenet,simo2018mastering,xu2019perceptual} and paint~\cite{huang2019learning,nakano2019neural, liu2021paint} with supervision from raster images. Other works focus on conditional line drawing generation given paired images, which are often collected for specific tasks ~\cite{Li_2019_CVPR,LIPS2019,yi2019apdrawinggan,artline2020}. In contrast, our method handles unpaired data and translates between sketches of different domains.

Our method is most similar to Unpaired Portrait Drawing Generation (UPDG)~\cite{yi2019apdrawinggan}, which creates portrait drawings from unpaired data. UPDG also uses an adversarial image translation setup, but modifies cycle-consistency for drawings, employs a truncation loss, and uses discriminators for the eyes, nose, and mouth. In contrast, our method is built on losses which encourage line drawings to carry meaningful information about geometry and semantics. Our objectives allow us to greatly reduce reliance on cycle consistency (or the appearance reconstruction), and to generate drawings for arbitrary photographs and not just portraits. 

Recent work has been successful at text-driven image editing and synthesis with the extensive shared visual-text embedding Contrastive Language-Image Pre-training (CLIP)~\cite{clip2021,vqganclip,patashnik2021styleclip}. CLIPDraw~\cite{frans2021clipdraw} also uses CLIP to create drawings, but with text inputs. This method requires no training, and simply minimizes the CLIP distance between a rasterized set of B\'ezier curves~\cite{Li:2020:DVG} and the text prompt. CLIPDraw demonstrates that the CLIP embedding can match semantics between text and drawings despite the domain gap. In contrast, previous methods have adapted new architectures to specifically examine semantics in line drawings~\cite{yu2017sketch, pixelor20siga}. Our approach similarly minimizes the distance between inputs and generated drawings in CLIP space, but instead conditions on an input photograph and generates drawings in multiple styles.

Our work also shares similarities to CyCADA~\cite{hoffman2018cycada} in that the output images are trained to semantically match the inputs. However, CyCADA applies this constraint with a pretrained classifier for a translation between source and target data for domain adaptation. In contrast, our semantic constraint makes use of the CLIP embedding, which can richly describe complex scenes.

Given two datasets, modern image translation and style transfer methods can transform images into new domains~\cite{gatys2016image, hertzmann2001image, isola2017image, johnson2016perceptual, zhu2017unpaired}. Modern approaches can produce high quality results given paired correspondences~\cite{isola2017image,wang2018high,esser2021taming,chen2017photographic}, however large aligned line drawing datasets are scarce. Fortunately, many approaches address image translation for unpaired data, often relying on an adversarial setup~\cite{zhu2017unpaired,yi2017dualgan,kim2017learning,chen2020reusing,Kim2020U-GAT-IT:,zhao2020unpaired,Shao_2021_ICCV,xie2021unaligned,anoosheh2018combogan,xie2021unaligned,nizan2020council}. Other methods translate images between domains by separating style and content~\cite{jiang2020tsit,huang2018munit,liu2019few}. Cheng et al. also use depth information to provide structure for neural style transfer~\cite{cheng2019structure}. Although these approaches are very successful at artistic style transfer and translating between rich domains with shape changes (e.g. dogs to cats, anime to selfies), they still generate create sparse line drawings which are missing key strokes.
\begin{figure}[t]
  \centering
  \includegraphics[width=\linewidth]{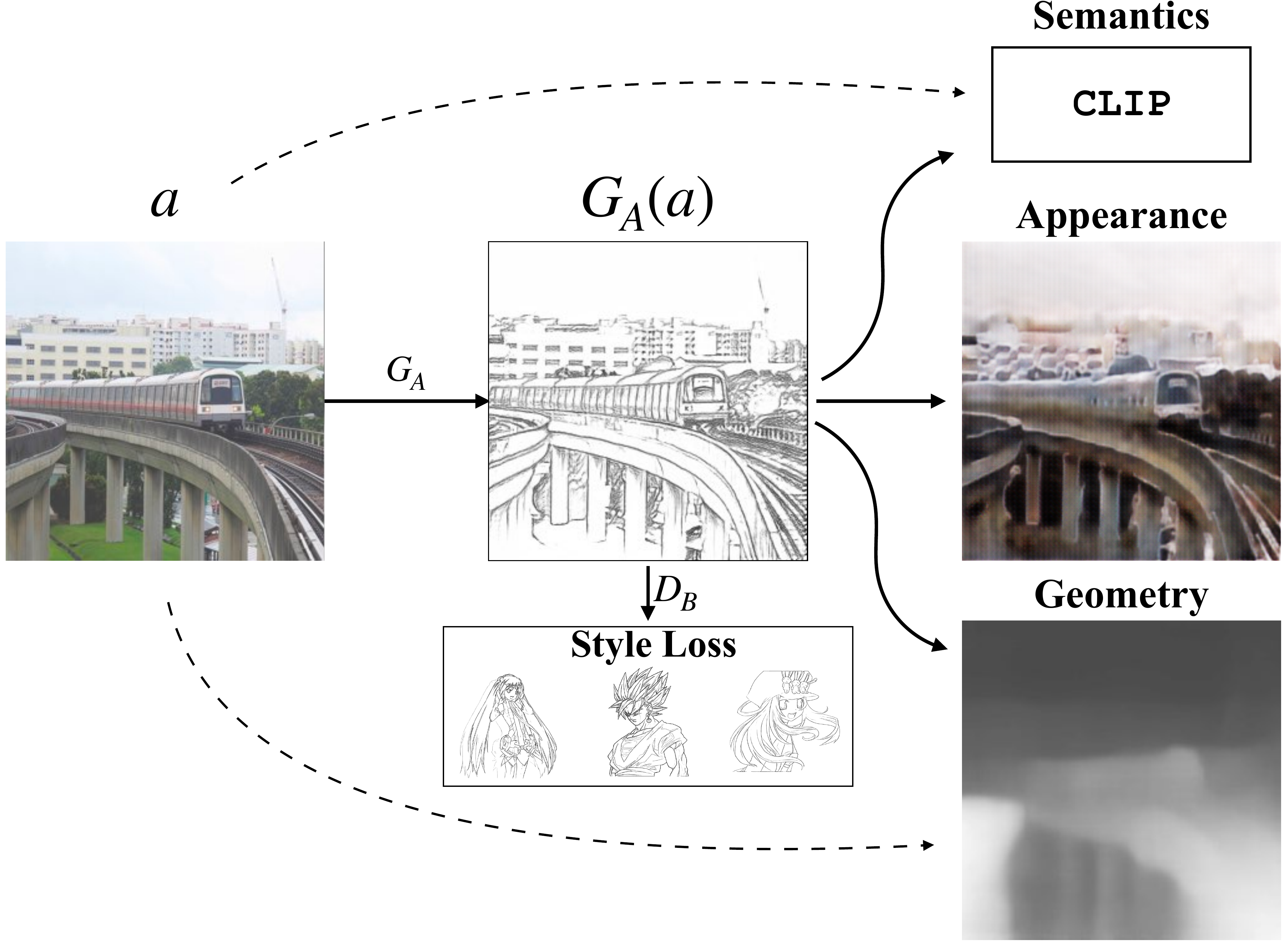}
  \caption{Given a photograph $a$, our model trains network $G_A$ to synthesize line drawing $G_A(a)$ via four main losses. Adversarial style loss with discriminator $D_B$ encourages generated line drawings to match the style of the training set. The \texttt{CLIP}, appearance, and geometry losses enforce that the line drawing communicates effective semantic, appearance, and geometry respectively.} \label{fig:system}
  \vspace{-0.4cm}
\end{figure}

\section{Method} \label{method}
Our goal is to train a model to automatically generate line drawings of arbitrary photographs given a dataset of photographs and an unpaired dataset of line drawings. We formulate this problem as unpaired image translation between domain $A$ which contains photographs, and domain $B$ which represents line drawings of a particular style. Most previous approaches solely consider preserving photographic appearance in the line drawing through cycle consistency. Instead, our method further directs this translation through objectives which assess the geometry and semantic information communicated by line drawings. This setup is shown in Figure~\ref{fig:system}. We show in Section~\ref{reesults} that these new losses are essential for creating meaningful drawings.

We use an adversarial training setup with generator networks $G_A$, $G_B$ and discriminators $D_A$, $D_B$ for domains $A$ and $B$ respectively. The geometry objective is implemented through a pretrained depth network which predicts depth maps from line drawings, and imposes a supervised loss on the depth outputs. This loss encourages our model to draw lines in geometrically important locations (e.g. occluding contours). Secondly, we introduce a CLIP~\cite{clip2021} loss to add semantic meaning into the generated line drawings. Because arbitrary photographs often display complex scenes, we use the visual CLIP embedding which captures semantic details quite well. We then impose that the CLIP embedding of the line drawing is similar to the CLIP embedding of the original photograph. We also use a weakly weighted cycle consistency loss to preserve appearance information.

\subsection{Losses}

\paragraph{The adversarial loss} encourages generated images to belong to their respective domains~\cite{goodfellow2014generative}. The loss for each domain using the LSGAN setup~\cite{mao2017least} is formulated below.
\begin{eqnarray} \label{eqn:ganlloss}
L_{GAN} =& \mathbb{E}_{a \sim A} [D_A(a)^2] + \mathbb{E}_{b \sim B} [\big( 1 - D_A(G_B(b)) \big)^2 ] \nonumber \\
 +& \mathbb{E}_{b \sim B} [D_B(b)^2] + \mathbb{E}_{a \sim A} [\big( 1 - D_B(G_A(a)) \big)^2 ] \nonumber \\
\end{eqnarray}

\paragraph{The geometry objective} maximizes depth information in generated line drawings during training. We observe that line drawings are often effective conveyors of 3D shape, and apply this property during training. Given a substantial dataset of line drawings, a model may learn this trait without any explicit supervision. However, current methods without such geometric constraints fail to place lines in meaningful places (see Section~\ref{reesults}). Domain gaps between the dataset of photographs and line drawings are also obstacles. Instead, we propose a geometric constraint which supervises depth predictions from line drawings.

To supervise depth predictions from line drawings, it is necessary to obtain depth maps for the photographic inputs. Unfortunately, ground truth depth information is usually unavailable for most datasets. However, recent methods are very successful at producing high resolution depth maps for photographs. This advance allows us to use pseudo-ground truth depth maps obtained from a state of the art depth prediction network $F$; in practice we use the network from \cite{Miangoleh2021Boosting}, which is based on MiDaS~\cite{Ranftl2019}. We note that pseudo-ground truth maps for photographs are only required for training, and not at test time.

A simple way to supervise geometry predictions would be to introduce network $G_{Geom}$ to predict depth maps from line drawings during training. However, this approach has several issues. Training $G_{Geom}$ to learn depth from synthetic line drawings may encourage line drawing generator $G_A$ to instill depth information in an unwanted form, such as an imperceptible signal~\cite{chu2017cyclegan}. We want to avoid accidentally embedding invisible information into our line drawings. Using pretrained depth network $F$ on line drawings is not an option because of the domain gap.

We propose instead to learn to infer depth from image features which are commonly shared between photographs and line drawings. Specifically, we pretrain a network $G_{Geom}$ to predict depth given ImageNet~\cite{deng2009imagenet} features. Such features, especially in early layers, are useful for transfer learning~\cite{kornblith2020s}. This scenario hopes to avoid the invisible signal issue by first encoding line drawings into a shared representation with photographs, and then applying a network which has learned depth from photographic features.

To obtain image features, we input photographs into pretrained Inception v3~\cite{szegedy2016rethinking} network and extract features from the Mixed 6b node (see supplemental).
We denote the extracted features at this layer for input $a$ as $I(a)$.
After pretraining, network $G_{Geom}$ provides depth map predictions for line drawings. In practice, we finetune $G_{Geom}$ while training line drawing generation.

The geometry loss is formulated below. Given photograph $a$, we first input $a$ into state of the art depth network $F$ and obtain pseudo-ground truth depth map $F(a)$. We then generate line drawing $G_A(a)$ and extract its ImageNet features $I(G_A(a))$.
These features are then passed to pretrained depth network $G_{Geom}$ to produce depth map prediction $G_{Geom}(I(G_A(a)))$. This depth prediction is then compared to the pseudo-ground truth depth map $F(a)$. Further details and depth reconstructions are in the supplementary.
\begin{eqnarray} \label{eqn:geom}
    L_{geom} &= &\| G_{Geom}(I(G_A(a))) - F(a) \| 
\end{eqnarray}
\paragraph{The semantics loss} is implemented by minimizing the distance between the CLIP embeddings of the input photograph and the generated line drawing. The goal of this objective is to convey semantic information from the original photograph into its corresponding synthesized line drawing. In computer vision, semantics are often learned in the form of labels and segmentation maps. However, these representations are limited in capacity to specific domains or objects. 
To encode semantic information from entire scenes, we use the shared visual-text embedding CLIP~\cite{clip2021}, which captures rich semantic information in both photographs and art~\cite{frans2021clipdraw,vqganclip}. We then penalize the distance in CLIP space between the generated line drawing and the original photograph. The objective is formulated below.
\begin{eqnarray} \label{eqn:CLIP}
    L_{\texttt{CLIP}} &= &\| \texttt{CLIP}(G_A(a)) - \texttt{CLIP}(a) \| 
\end{eqnarray}
\paragraph{The appearance loss} (or cycle consistency) has been used to encode input appearance through image translation~\cite{zhu2017unpaired,kim2017learning}. The appearance loss for each direction of the mapping is below.
\vspace{-0.1cm}
\begin{eqnarray} \label{eqn:cycle}
L_{cycle} = \| G_B(G_A(a)) - a \| + \| G_A(G_B(b)) - b \|
\end{eqnarray}
\subsection{Full Objective}
Our full objective is:
\begin{eqnarray} \label{eqn:full}
L = & \lambda_{\texttt{CLIP}}L_{\texttt{CLIP}} + \lambda_{geom}L_{geom} \nonumber \\
&+ \lambda_{GAN}L_{GAN} + \lambda_{cycle}L_{cycle}
\end{eqnarray}
In practice we set $\lambda_{\texttt{CLIP}}=10$, $\lambda_{geom}=10$, $\lambda_{GAN}=1$, $\lambda_{cycle}=0.1$.

\paragraph{Implementation} We use an encoder-decoder generator architecture with Res-Net blocks in the middle~\cite{johnson2016perceptual,zhu2017unpaired,he2016deep}, and a patch-based discriminator~\cite{isola2017image}. The architecture for pretrained depth network $G_{Geom}$ is based on the Global Generator from pix2pixHD~\cite{wang2018high} and further detailed in the supplemental material. We use MSE error for the CLIP loss and $L1$ distance for the appearance and geometry losses. We use Adam~\cite{adamopt} to optimize with a learning rate of $0.0002$ and train for at least $30$ epochs with batch size $6$.
\section{Experiments}\label{reesults}

\begin{figure*}[h]
  \centering
  \includegraphics[width=\linewidth]{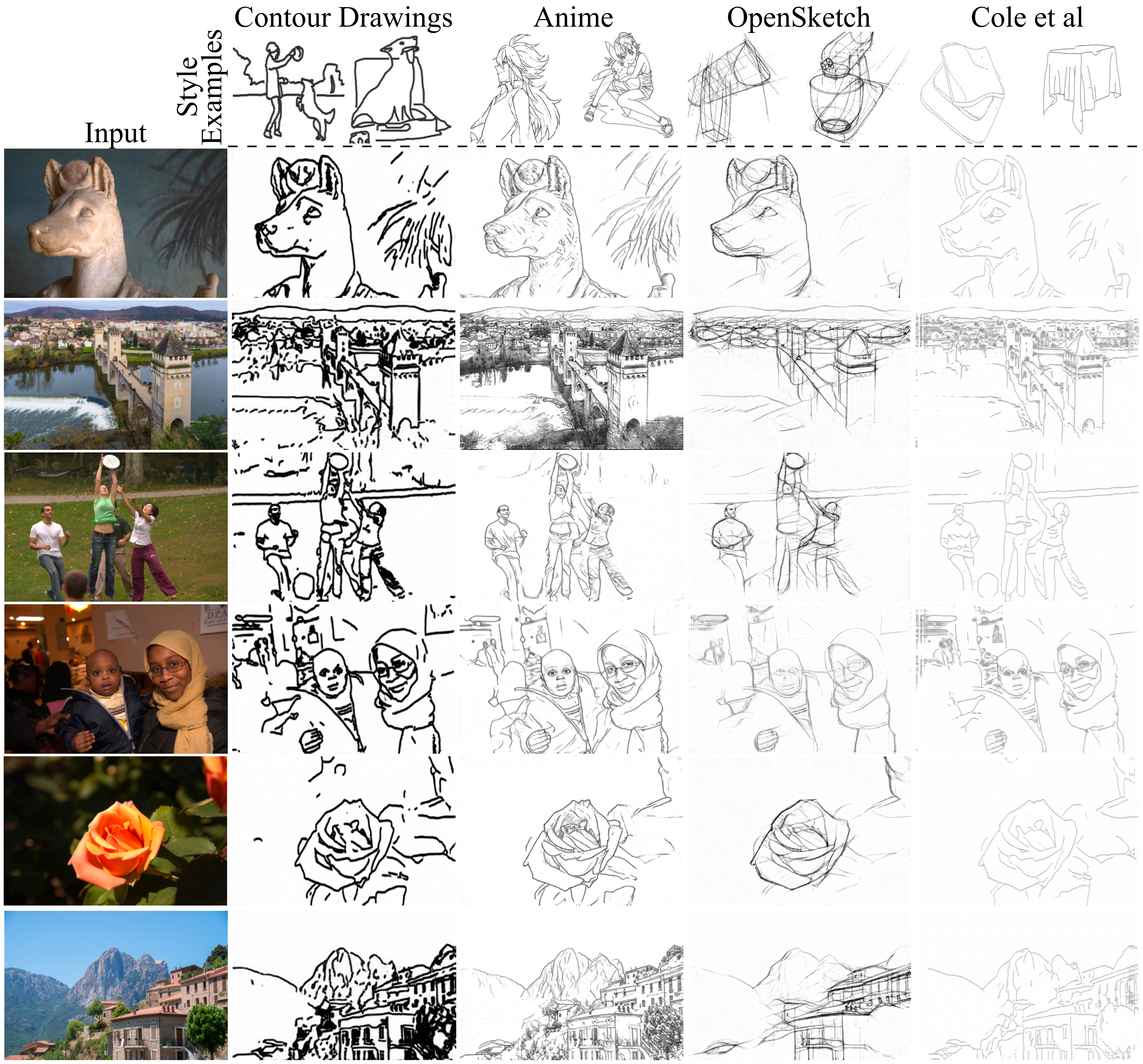}
  \caption{Results of our method in four different styles.} \label{fig:styles}
  \vspace{-0.3cm}
\end{figure*}

We evaluate our described approach and provide qualitative and quantitative comparisons for both general photographs and portraits in multiple styles.

\subsection{Line Drawings from Photographs} \label{photo_exp}

Our first evaluation task is to generate line drawings from photographs of arbitrary scenes. Below we describe the datasets for training and evaluation.

\paragraph{Datasets} For training, our method requires a dataset of photographs and a separate dataset of line drawings. We train on a randomly selected $10,000$ image subset of the Common Objects in Context (COCO)~\cite{lin2014microsoft} dataset which contains a variety of scenes. For evaluation, we create line drawings from photographs in the MIT-Adobe FiveK dataset~\cite{bychkovsky2011learning}. This dataset contains high quality images of many subjects (landscapes, buildings, people, etc).

We train multiple models with different styles of line drawings. Examples for each style are shown in Figure~\ref{fig:styles}. Quantitative evaluations are performed for two styles of line drawings: \textbf{1) The Contour Drawings dataset~\cite{LIPS2019}} contains $5,000$ drawings for various scenes (often with humans or dogs). \textbf{2) The Anime Colorization dataset~\cite{animedataset2020}} consists of $14,224$ sketches of various anime characters. Qualitative results in the style of OpenSketch~\cite{gryaditskaya2019opensketch} and artist drawings from Cole et al.~\cite{cole2008people} are shown in Figure~\ref{fig:styles}.

\begin{figure*}
  \centering
  \includegraphics[width=\linewidth]{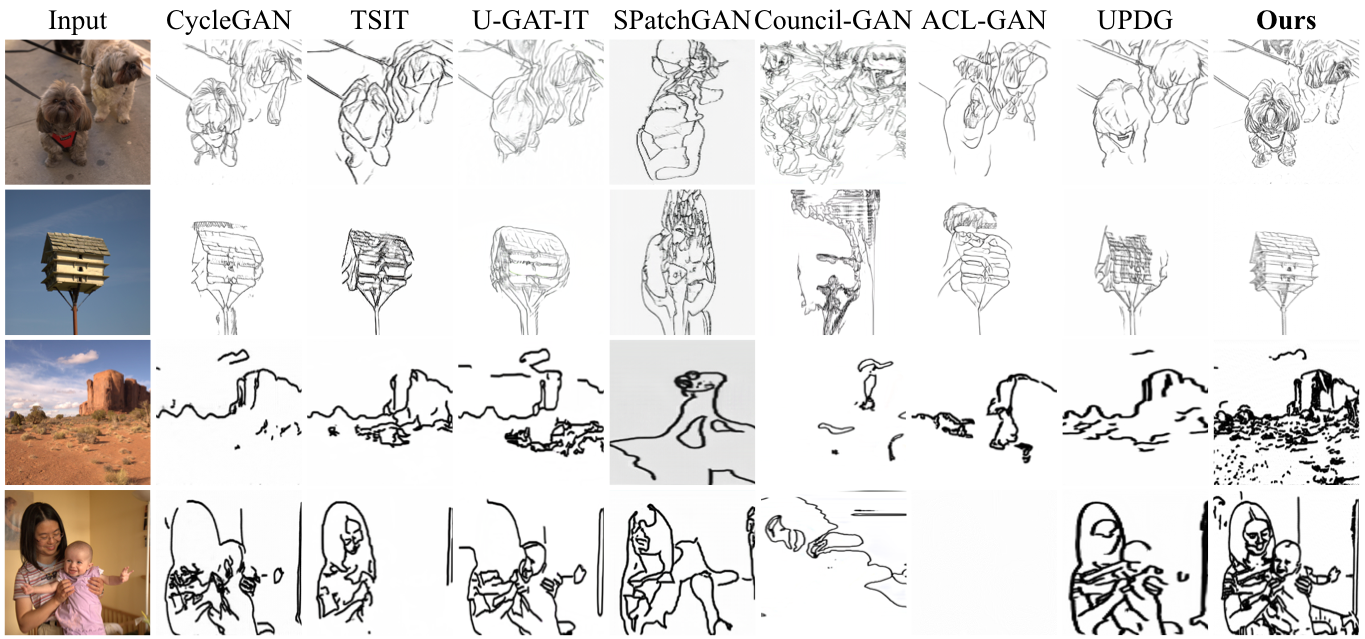}
  \caption{Comparison with other methods. \emph{Left to right:} Input photograph, CycleGAN, TSIT, U-GAT-IT, SPatchGAN, Council-GAN, ACL-GAN, UPDG, and Our approach. All methods are trained using the same data on two styles of line drawings. Our method produces the most detailed drawings capturing important aspects of the original photograph.} \label{otherphoto} \vspace{-0.2cm}
\end{figure*}
\paragraph{Comparison methods} We compare our approach to state-of-the-art unpaired image-to-image translation methods for the photograph to line drawing task. These methods include: \textbf{1) CycleGAN~\cite{zhu2017unpaired}} uses an appearance loss and a patch-based discriminator~\cite{isola2017image}. \textbf{2) TSIT~\cite{jiang2020tsit}} creates images by combining features from separate content and style streams. \textbf{3) U-GAT-IT~\cite{Kim2020U-GAT-IT:}} uses an attention module and auxiliary classifier and cycle consistency. \textbf{4) ACL-GAN~\cite{zhao2020unpaired}} relaxes strict pixel cycle consistency into distributional level consistency \textbf{5) Unpaired Portrait Drawing Generation (UPDG)~\cite{YiLLR20}} creates line drawings in multiple styles for portrait drawings. This method builds upon CycleGAN with discriminators for facial features, a truncation loss, and a modified cycle loss using HED images~\cite{xie2015holistically}. For the photograph task, we do not include the face discriminators as they do not apply to arbitrary photographs without human subjects. We also provide qualitative comparisons with SPatchGAN~\cite{Shao_2021_ICCV} and Council-GAN~\cite{nizan2020council} in Figure~\ref{otherphoto}.

\paragraph{Qualitative comparison} Figure~\ref{otherphoto} shows compares our method to previous work in two styles. Other methods commonly fail to place lines in meaningful locations, whereas our drawings have recognizable features and boundaries. Some methods such as SPatchGAN, Council-GAN, and ACL-GAN attempt to strictly stay close to the training set domain. This is most noticeable for the Anime style, as these approaches often produce drawings which resemble anime characters over the input photographs.

\paragraph{User Study} We conduct a user study to perceptually compare our approach with other methods. In this study, participants were shown a reference photograph, and two line drawings of the same photograph made by different methods. Users were then asked to select the line drawing that best depicts the input photograph. For this study we showed users up to $100$ images and there were $184$ unique participants. $1000$ judgments were made for each comparison. Table~\ref{table:mturkMethods} reports the percentage users chose line drawings from our method over various baselines. Users overwhelmingly preferred line drawings created by our method in all cases.

\begin{table}[t]
  \centering
  \resizebox{0.8\linewidth}{!}{
  \begin{tabular}{cccc}
    \toprule
      & Contour Drawings & Anime & Total  \\
\midrule
CycleGAN\cite{zhu2017unpaired}        &     $98.7$\%  &     $87.3$\%  &     $93$  \%  \\ \hline
TSIT\cite{jiang2020tsit}              &     $99.6$\%  &     $95.3$\%  &     $97.5$\%  \\ \hline
U-GAT-IT\cite{Kim2020U-GAT-IT:}       &     $99.5$\%  &     $97.3$\%  &     $98.4$\%  \\ \hline
ACL-GAN\cite{zhao2020unpaired}        &     $100$\%   &     $97.5$\%  &     $98.8$\% \\ \hline
UPDG\cite{YiLLR20}                    &     $98.9$\%  &     $96.7$\%  &     $97.8$\%  \\

    \bottomrule
  \end{tabular}
  }
  \caption{User study results comparing to different unpaired translation methods. We report the percentage of times users preferred our approach over the other methods.} \label{table:mturkMethods} \vspace{-0.8cm}
\end{table}

\paragraph{Ablation Study} We perform an ablation study to verify the inclusion of each loss. Three versions of our model are trained: without the geometry loss, without the CLIP loss, and without the appearance or cycle loss. We compare each ablation to our full method. We use the perceptual study setup described above and report the percentages users selected our full method over each ablation in Table~\ref{table:mturkAblations}. The CLIP loss was essential for all styles, while the Contour Drawings style relies on the depth loss much more than the Anime style. The appearance loss improves results slightly. 

\begin{figure*}[t]
  \centering
  \includegraphics[width=\linewidth]{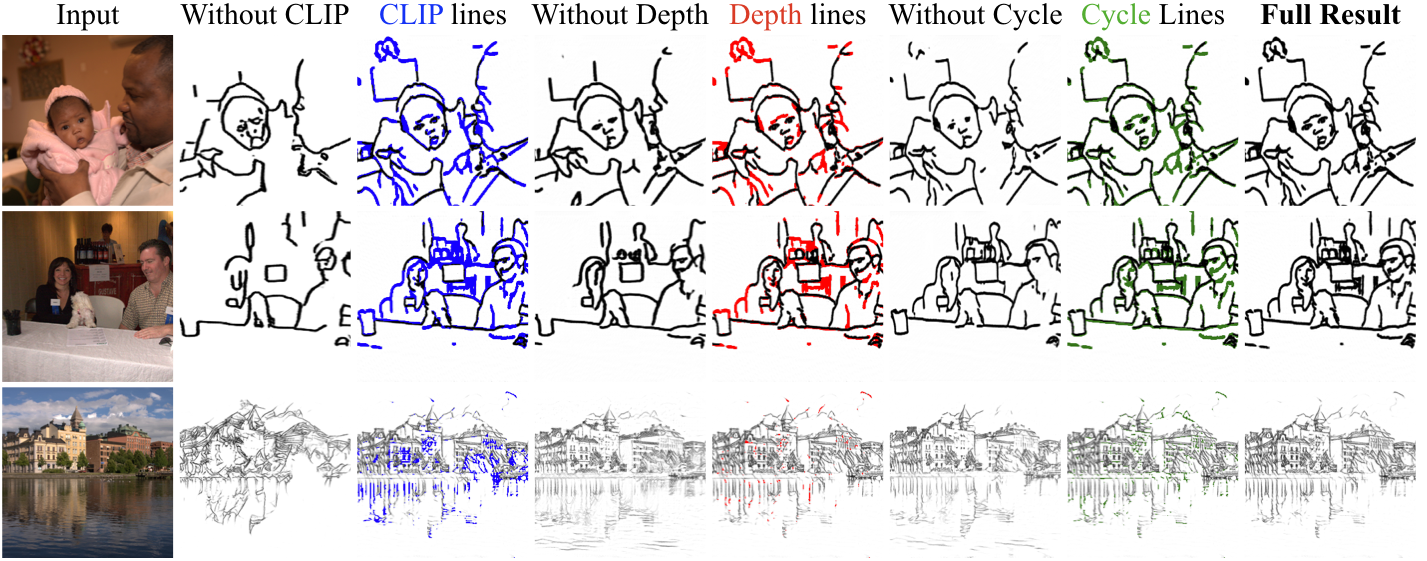}
  \caption{Ablations of our method, and our full result. For each ablations, we show the lines added to get the full result by including each loss. These lines are in blue for CLIP, red for depth, and green for appearance. The CLIP loss adds the most lines, while the depth loss adds more information and occluding contours in the second row. The appearance loss adds small strokes and shading for the Anime style.} \label{fig:ablations} \vspace{-0.3cm}
\end{figure*}

Figure~\ref{fig:ablations} shows qualitative examples from all ablations. The CLIP loss adds the most lines. In some cases, styles with a high density of lines may totally rely on the CLIP loss. We find this situation to be the case for the Anime style, whose `without depth' ablation is comparable to the full method. The depth loss is most useful for sparse styles such as the Contour Drawings style, where it adds occluding contours and textures. We note that the semantic loss improves geometry, and depth information can help semantics as well. The cycle loss improves result quality by preserving appearance aspects such as textures and outlines. However, removing the cycle loss does not qualitatively affect results significantly.

\begin{table}[t]
  \centering
  \resizebox{0.9\linewidth}{!}{
  \begin{tabular}{cccc}
    \toprule
      & Contour Drawings & Anime & Total  \\
\midrule
Without depth    &     $92.2$\%  &     $48.3$\%  &     $70.3$\%  \\ \hline
Without CLIP     &     $98.9$\%  &     $84.9$\%  &     $92$\% \\ \hline
Without Cycle Consistency    &     $87.0$\%  &     $64.9$\%  &     $76$\%  \\
    \bottomrule
  \end{tabular}
  }
  \caption{User study results for the ablation study. We report the percentage users chose the full method over the ablations.} \label{table:mturkAblations}
  \vspace{-0.2cm}
\end{table}

\begin{table}[t]
  \centering
  \resizebox{0.84\linewidth}{!}{
  \begin{tabular}{cccc}
    \toprule
      & Contour Drawings & Anime & Total  \\
\midrule
CycleGAN      &     $58.0$\%  &     $65.1$\%  &     $62.0$\%  \\ \hline
Ours          &     $68.4$\%  &     $66.8$\%  &     $67.6$\% \\ \hline
Photograph    &     $-$       &     $-$       &     $70.3$\%     \\
    \bottomrule
  \end{tabular}
  }
  \caption{User study results for relative depth prediction. We report the percentage of times users chose the closer point correctly for each baseline. For both styles, users correctly inferred relative depth more often in drawings from our method over CycleGAN.} \label{table:geom} \vspace{-0.5cm}
\end{table}

\begin{table}[t]
  \centering
  \resizebox{0.92\linewidth}{!}{
  \begin{tabular}{ccccc}
    \toprule
      & Contour Drawings & Anime & Total & Unrecognizable  \\
\midrule
CycleGAN      &     $0.7436$  &    $0.8074$   &     $0.7799$  &    $26.7$\% \\ \hline
Ours          &     $0.8160$  &    $0.8371$   &     $0.8274$  &    $13.7$\% \\ \hline
Photograph    &     $-$       &    $-$        &     $0.8804$  &    $0.02$\% \\ 
    \bottomrule
  \end{tabular}
  }
  \caption{Mean cosine similarity between captions describing line drawings and captions describing the input photographs. The last column reports the percentage of images that users could not identify. Our line drawings are more easily described and recognizable.} \label{table:semantics} \vspace{-0.6cm}
\end{table}

\paragraph{Evaluating Geometry and Semantics in Drawings}\label{geomandsemanticsexp}
We design two experiments to evaluate the depth and semantic information conveyed in the generated line drawings. To examine depth information, we conduct a user study to assess if humans can correctly infer relative depth from our drawings. Participants viewed an image with two randomly placed points and were asked to identify the point closest to the camera, similarly to \cite{chen2016single}. We perform this evaluation on drawings from our method, CycleGAN, and on photographs. Table~\ref{table:geom} reports the percentage each baseline agreed with the pseudo-ground truth depth predictions. In general, users inferred the correct relative depth more often in our drawings, especially for the Contour Drawings style. For the Anime style, relative depth predictions were better for our results by a slim margin. This result complements the ablation study, where the depth loss was not as effective for the Anime style. If relative depth can already be inferred from CycleGAN (despite lower drawing quality), then the geometry objective may not have much impact. In contrast, the depth loss greatly improves both relative depth predictions and drawing quality for the Contour Drawing style.

To assess semantic meaning, we show users a photograph and ask them to write a one sentence caption for the image. Participants were also given the option to designate images as unrecognizable. Users viewed results from our method, CycleGAN, and photographs. Each caption is encoded in CLIP space and then compared to the mean CLIP embedded photograph caption using cosine similarity. Table~\ref{table:semantics} reports the mean cosine similarities and the percentage of unrecognizable images. In all cases our method produces more accurate descriptions and recognizable drawings.

\subsection{Line Drawings from Portraits}
While our method was not designed specifically for portraits, we compare to methods specialized for this task. We use two main settings for comparison. Firstly, we compare to other methods directly on styles they present. Then we provide a second comparison where we train our model on unpaired portraits from the Helen Facial Feature Dataset~\cite{le2012interactive} in the style of the APDrawings dataset~\cite{yi2019apdrawinggan}. Details for each dataset are provided in the supplemental.

\begin{figure*}
  \centering
  \includegraphics[width=\linewidth]{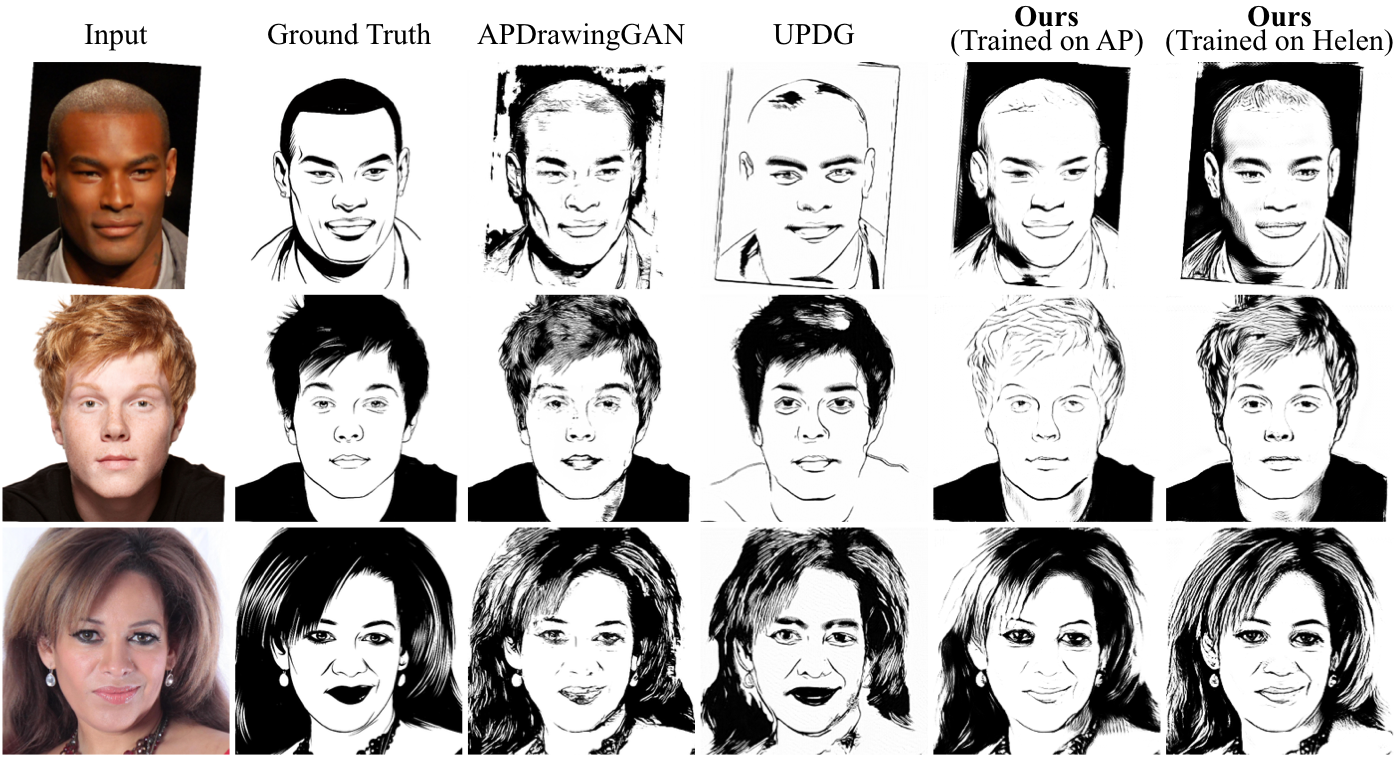}
  \caption{Results for several methods on APDrawings test data. \emph{Left to right:} Portrait photograph, artist's drawing, APDrawingGAN, UPDG (trained on Helen), Our result (trained on APDrawings), our result (trained on Helen). All methods were trained with the APDrawings line art style. Our approach produces accurate and well formed drawings.} \label{fig:portraits1} \vspace{-0.3cm}
\end{figure*}

\paragraph{Comparisons} \textbf{1) APDrawingGAN~\cite{yi2019apdrawinggan}} uses supervised adversarial training to create line drawings in the style of the paired APDrawings. In one comparison, we train our model on APDrawings directly. This setting disadvantages our method because we do not use paired supervision. However, our method can use unpaired data and we exploit this property in the next case. We then use portraits from the Helen dataset to train a separate model, while keeping the drawing style of APDrawings. Our second comparison evaluates our method trained on the Helen dataset against supervised APDrawingGAN results.

\textbf{2) Unpaired Portrait Drawing Generation (UPDG)~\cite{YiLLR20}} is described in Section~\ref{photo_exp}. In the first setting, we compare to a pretrained UPDG model in the style of illustrators Charles Burns~\cite{charlesburns} and Yann Legendre~\cite{yannlegendre} (style 1 from~\cite{YiLLR20}). We train our model from scratch on an approximation of these datasets (see supplemental), and evaluate on the Helen test set. Secondly, we train both our approach and UPDG from scratch to create portraits from the Helen dataset in the style of APDrawings. We then compare on test portraits from APDrawings.

\paragraph{Qualitative comparison} Figure~\ref{fig:portraits1} shows portrait drawings created with APDrawings from all methods. APDrawingGAN produces reasonable results, while UPDG struggles with the line art style. We achieve decent results training on APDrawings, but quality drastically improves by training on the Helen dataset. Both our method and UPDG create high quality drawings in style 1 (see supplemental).

\paragraph{User Study} We perform a user study for all portrait comparisons. Participants were shown a portrait and two line drawings from different methods and asked to select the drawing which best depicts the subject in the portrait. Table~\ref{table:mturkMethodsPortrait} reports the percentage of times users chose our approach over the baselines. In case $1$, users preferred the supervised APDrawingGAN over our method (trained on APDrawings), but found our method (trained on Helen) preferable or comparable in case $2$. In general, UPDG struggles with the APDrawings style, and overall users slightly preferred our method for style 1.

\begin{table}[t]
  \centering
  \resizebox{0.64\linewidth}{!}{
  \begin{tabular}{ccc}
    \toprule
      & Case 1 & Case 2  \\
\midrule
APDrawingGAN\cite{yi2019apdrawinggan}    &     $36.7$\%  &     $60.1$\%   \\ \hline
UPDG\cite{YiLLR20}                       &     $64.2$\%  &     $94.8$\%   \\

    \bottomrule
  \end{tabular}
  }
  \caption{Perceptual study results for portrait comparisons. We report the percentage users chose our approach over each baseline. Case 1 compares both baselines on their datasets and styles. In case 2, we train our model on Helen in the style of APDrawings and compare to baselines trained on the same style.} \label{table:mturkMethodsPortrait}
  \vspace{-0.5cm}
\end{table}
\section{Discussion} \label{discussion}
% \vspace{-0.2cm}
\paragraph{Loss Formulations}
We explored several variants of the geometry and semantic losses in initial experiments. This includes using normal maps and multi-view consistency. We found the normal maps helpful for 3D shapes, however normal estimates are often noisy for photographs. Novel view prediction and using other 3D approaches are directions we hope to explore in future work. We selected depth prediction~\cite{Miangoleh2021Boosting} due to its robustness on photographs, and because we can reliably obtain depth predictions from image features that also can be extracted from line drawings. For the semantic loss, we explored finetuning image classifiers and segmentation networks on drawings and comparing intermediate features from these networks~\cite{deng2009imagenet,chen2017deeplab}. For a visual comparison, see the supplemental material.

\paragraph{Limitations} Our method is built on some limiting assumptions. We rely on pseudo-ground truth depth maps from a pretrained network for geometry supervision. Because we essentially distill this pretrained depth prediction network, our model has similar failure cases and biases.

Our model produces meaningful line drawings for many styles, but has failure cases shown in the supplemental. Our method is based on the hypothesis that a good line drawing accurately conveys depth and semantics, however some styles focus on the essence of the scene and not precision. We also struggle with certain lighting conditions and textures. Overall, the CLIP loss drives results to look more `photographic,' which may or may not be desirable. In some cases, this causes results to converge to grayscale photos.

\paragraph{Negative Impacts}

As with most data-driven techniques, our approach can learn bias in training. For instance, the Anime sketch dataset in Section~\ref{reesults} contains drawings of mostly feminine subjects. In addition, artistic datasets (such as the full Anime dataset used for creating line drawings) may contain sensitive content (e.g. nudity, weapons) whose influence could be visible in the output. 

\paragraph{Conclusion} Our approach creates compelling line drawings given unpaired data. This paper views line drawings as encodings of geometry, semantics, and appearance from real scenes. We built these ideas into a method which explicitly evaluates these properties through depth prediction, CLIP features, and image reconstruction to create line drawings from photographs.

\paragraph{Acknowledgements} We would like to thank Hyojin Bahng for proofreading the paper. This work was partially supported by a Packard Fellowship to PI, and the National Science Foundation under Grant No. 2105819.

%%%%%%%%% REFERENCES
{\small
\bibliographystyle{ieee_fullname}
\bibliography{main}
}

\clearpage


%%%%%%%%% BODY TEXT

\section{Supplemental Material}

\begin{figure}[h]
  \centering
  \includegraphics[width=\linewidth]{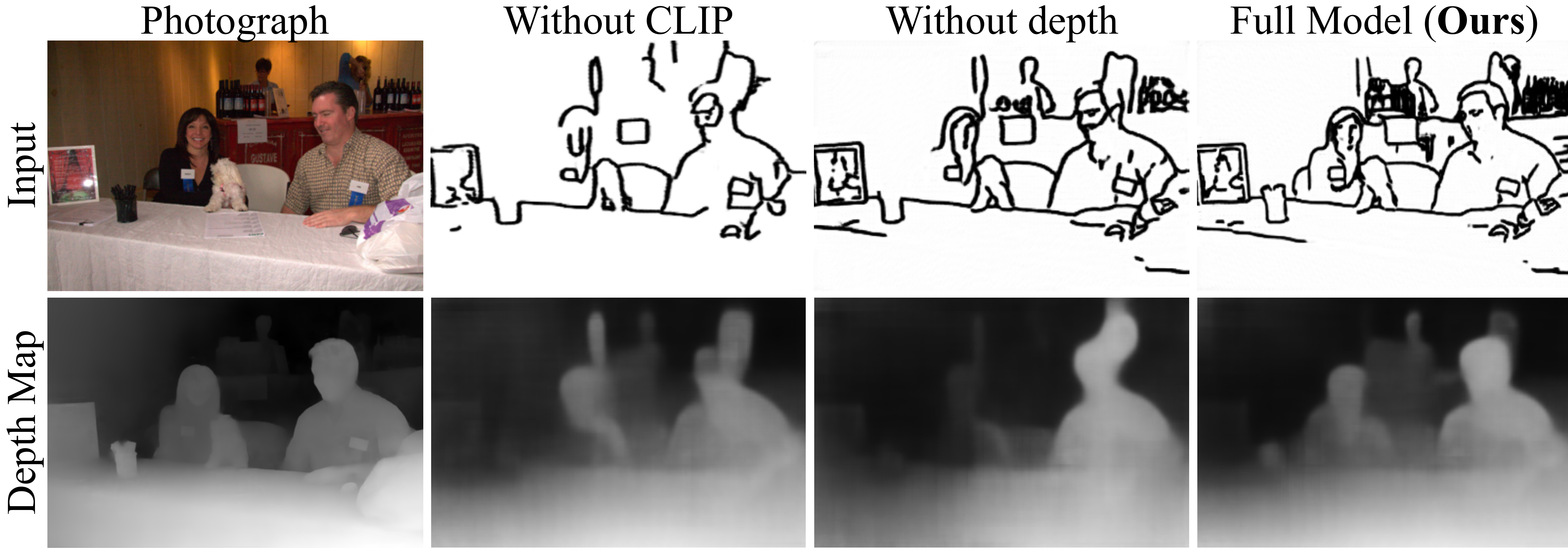}
  \vspace{-0.3cm}
  \caption{Depth predictions for various inputs. From left to right, we show the input photograph, the ablation trained without the semantic loss, the ablation trained without the depth loss, and our full model. The bottom row shows depth maps predicted from the images on the top row.} \label{fig:depthpreds}
  \vspace{-0.4cm}
\end{figure}

\subsection{More Results}

\begin{figure*}
  \centering
  \vspace{-0.4cm}
  \includegraphics[width=\linewidth]{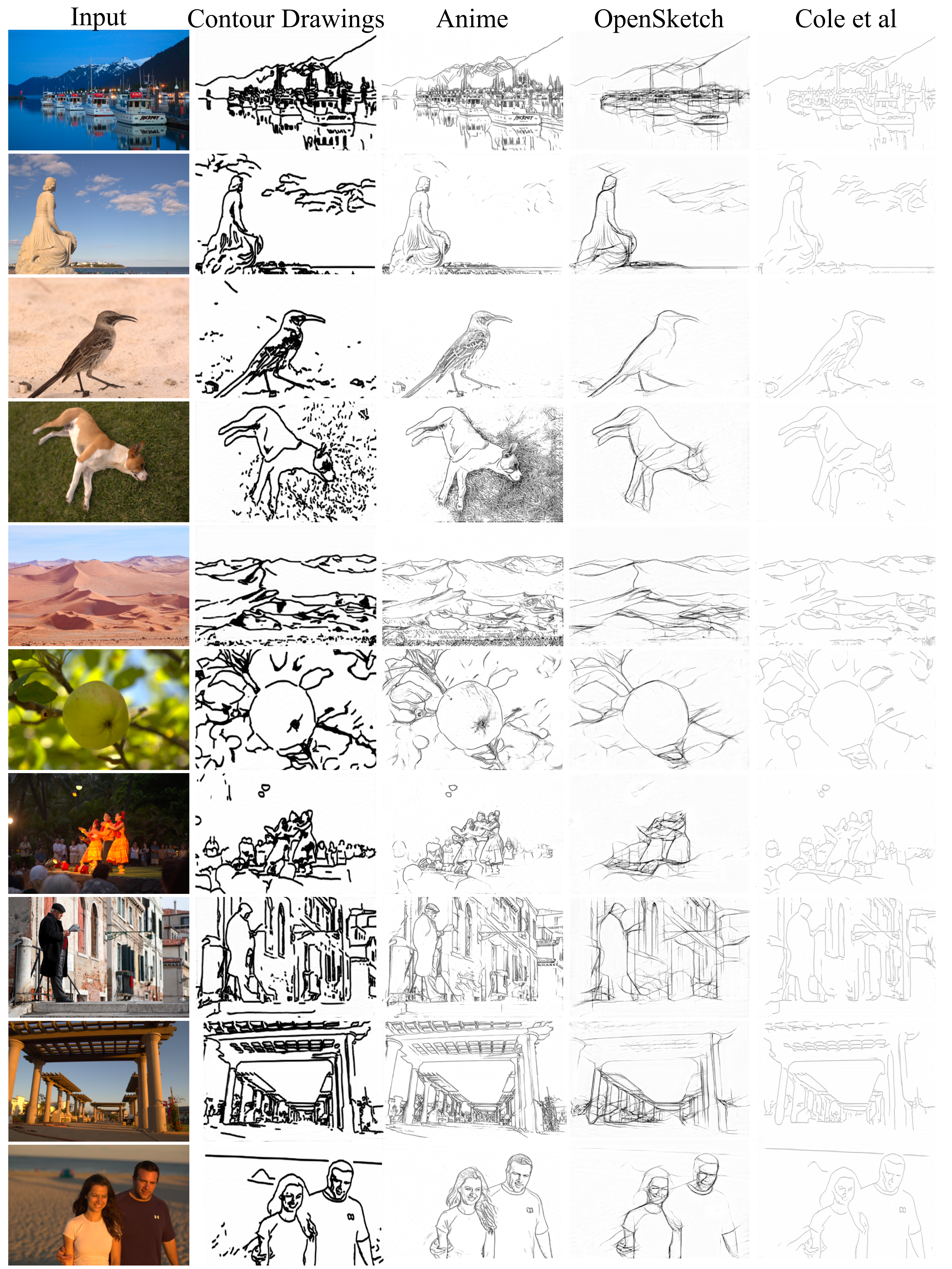}
  \caption{More results of our method in four different styles.} \label{fig:styles2}
  \vspace{-0.3cm}
\end{figure*}

In this section we provide more results. We show further results on four styles of line drawings in Figure~\ref{fig:styles2}. The OpenSketch and Cole et al tend to focus on the center of the image and sometimes are missing lines.

\subsubsection{Depth Reconstruction}

\begin{figure*}[h]
  \centering
  \includegraphics[width=\linewidth]{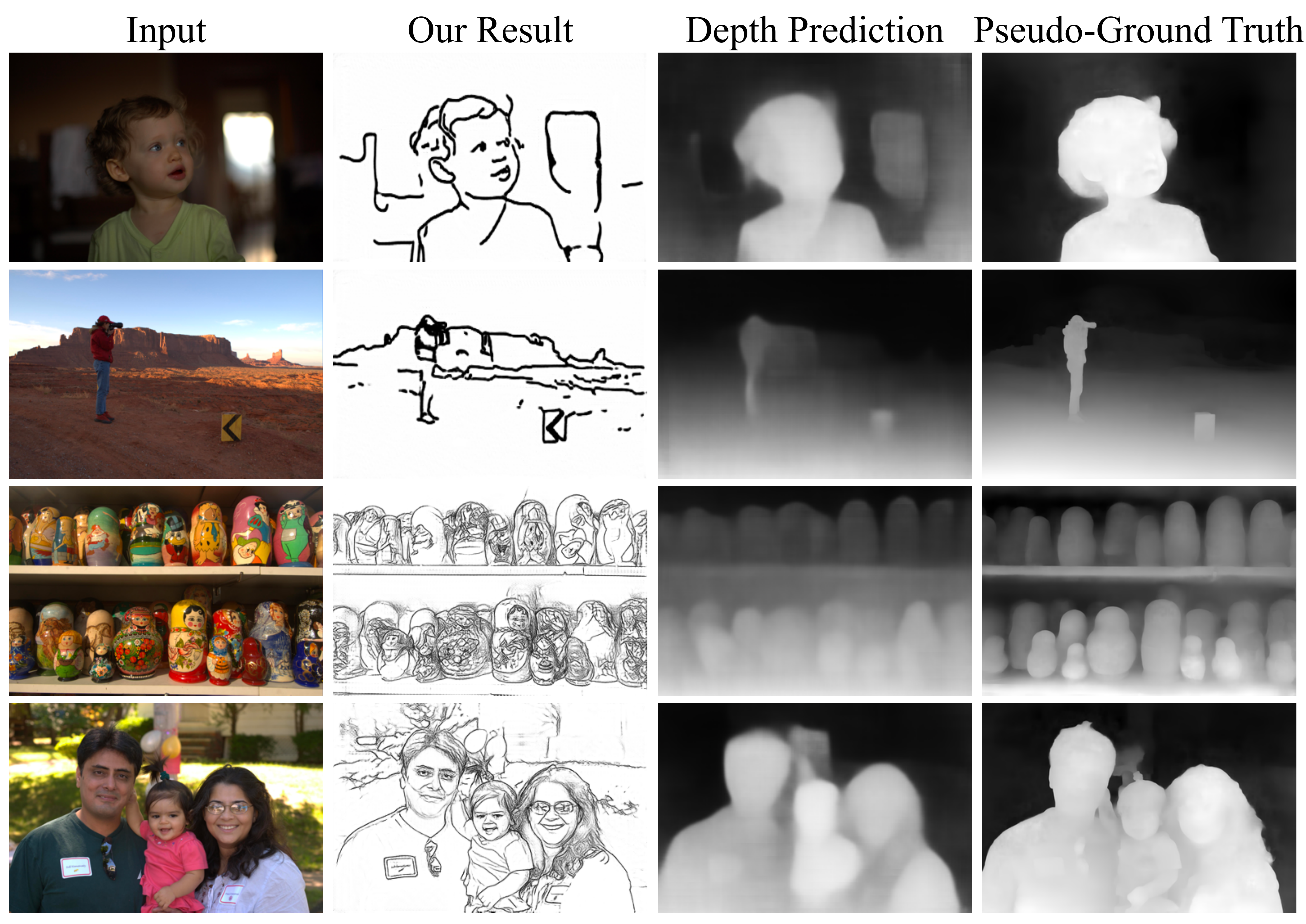}
  \caption{Depth predictions for our line drawings in two different styles. \emph{Left to right:} Input photograph, our line drawing result, the predicted depth map from the line drawing from $G_{Geom}$, and the pseudo-ground truth depth map from pretrained network $F$.} \label{fig:depthpred}
  \vspace{-0.3cm}
\end{figure*}

We show depth predictions from our line drawings in Figure~\ref{fig:depthpred} and compare to the pseudo-ground truth. Our depth prediction maps are coarse, but often capture key elements and relative depth of a scene. However, they are not always accurate as seen in the first and second rows. Our depth network sometimes interprets boundary lines as close objects rather than part of the background (such as the doorway in the first row of Figure~\ref{fig:depthpred}). Additionally, we find depth information is not so useful for scenes where all objects are either very close or very far from the camera. Adding the semantic loss also increases predicted depth map quality.

We provide a comparison of the depth maps predicted from line drawings in Table~\ref{table:depthError} and report the mean squared error for different ablations. Depth maps from state of the art pretrained model~\cite{Miangoleh2021Boosting} are used as a pseudo ground truth. We find that the reported errors are consistent with our results - adding the depth loss improves depth maps for the contour drawing style, whereas the anime style already portrays depth information well without the depth loss. Figure~\ref{fig:depthpreds} shows qualitative results and pseudo ground truth depth maps.

\begin{table}[h]
  \centering
  \resizebox{0.9\linewidth}{!}{
  \begin{tabular}{cccc}
    \toprule
      & Contour Drawings & Anime & Total  \\
\midrule
Without depth    &     $0.0530$  &     $0.0400$  &     $0.0465$  \\ \hline
Full model (with depth)     &     $0.0506$  &     $0.0418$  &     $0.0462$ \\ \hline
    \bottomrule
  \end{tabular}
  }
  \vspace{-0.1cm}
  \caption{Predicted depth map MSE errors for different ablations of our model.} \label{table:depthError}
  \vspace{-0.6cm}
\end{table}

\begin{figure}[h]
  \centering
  \includegraphics[width=\linewidth]{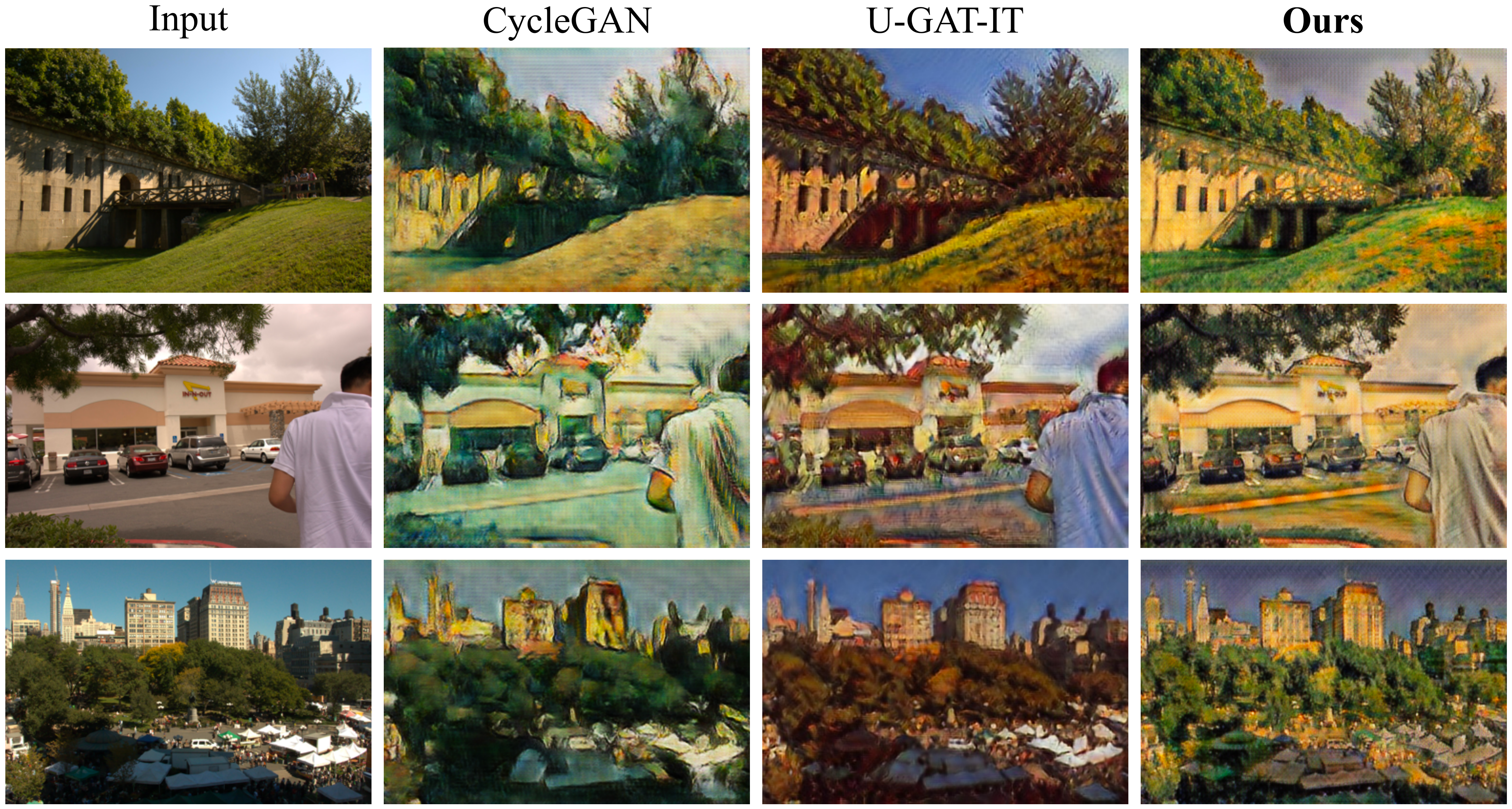}
  \caption{Our method designed for line drawing generation using Cezanne paintings as the style produces results comparable with other methods. The geometry and semantics loss works best with sparser styles. \emph{Left to right:} input photograph, CycleGAN results, U-GAT-IT results, and our method. Results are overall comparable.} \label{fig:cezanne}
  \vspace{-0.3cm}
\end{figure}

\subsubsection{Painterly styles}

Although our method is specifically designed for line drawings where the geometry and semantics increase the quality of sparser styles, we try creating images in the style of Cezanne paintings. Results are seen in Figure~\ref{fig:cezanne} where we provide comparisons to CycleGAN~\cite{zhu2017unpaired} and U-GAT-IT~\cite{Kim2020U-GAT-IT:}. All approaches handle the Cezanne style well and produce comparable results.

\subsubsection{Portraits in Style 1}

\begin{figure}[h]
  \centering
  \includegraphics[width=\linewidth]{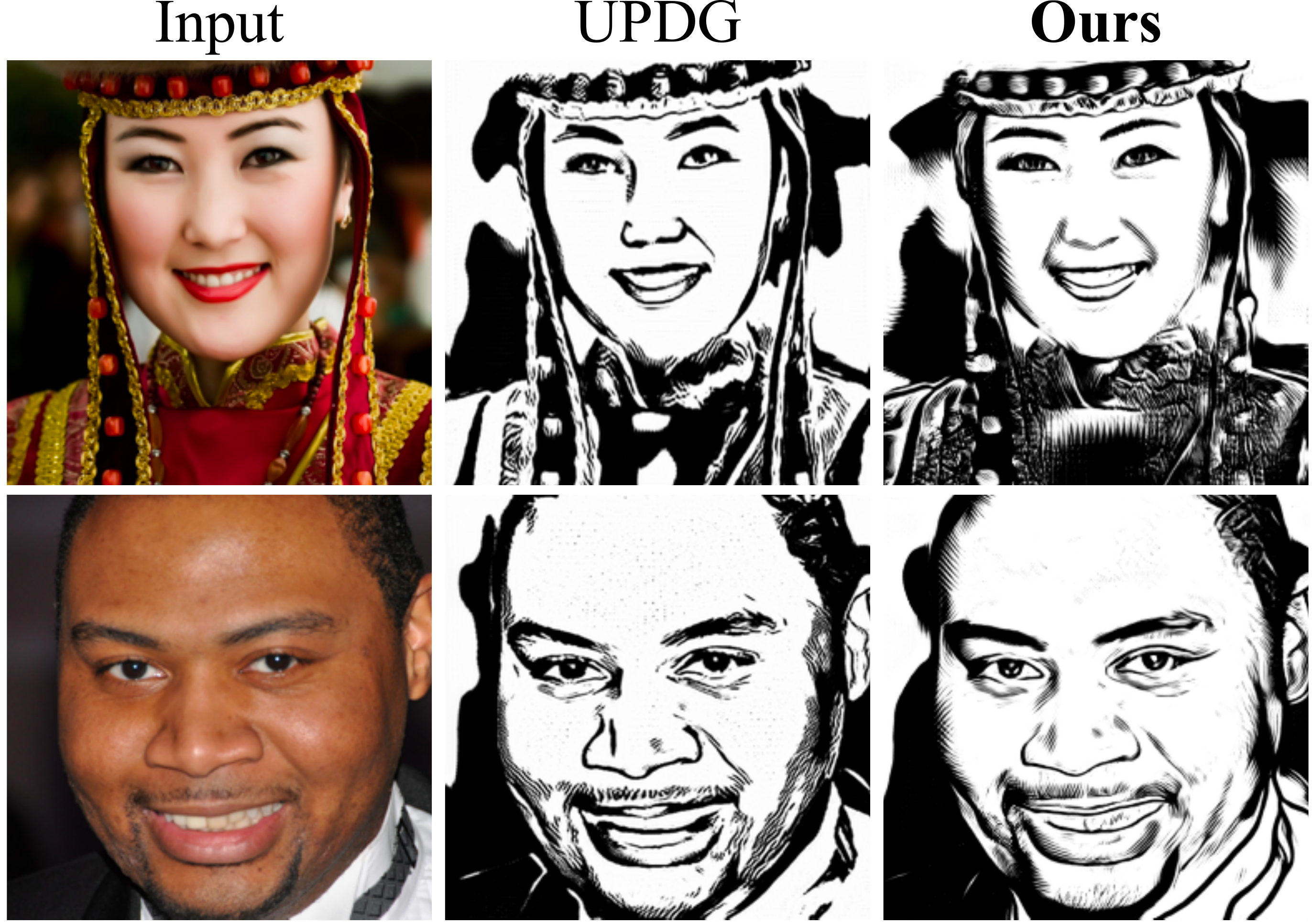}
  \caption{Comparison between UPDG results on style 1 from~\cite{YiLLR20}. Our method can be applied to arbitrary photographs whereas UPDG specifically creates portraits. \emph{Left to right:} Portrait photograph, UPDG, and our method. Although we were unable to exactly match the training data, both methods were trained using portraits `in the wild' and using line art style 1 from~\cite{YiLLR20}. Both approaches often produce nice drawings of the subject.} \label{fig:style1}
\end{figure}

We provide further comparisons for methods which specifically create portrait drawings, even though our method is designed for arbitrary photographs. Figure~\ref{fig:style1} provides visual examples for the comparison to UPDG in style 1 from their paper~\cite{YiLLR20}. Overall, both methods create nice portraits.

\subsection{Other loss variants}

\begin{figure}[h]
  \centering
  \vspace{-0.2cm}
  \includegraphics[width=\linewidth]{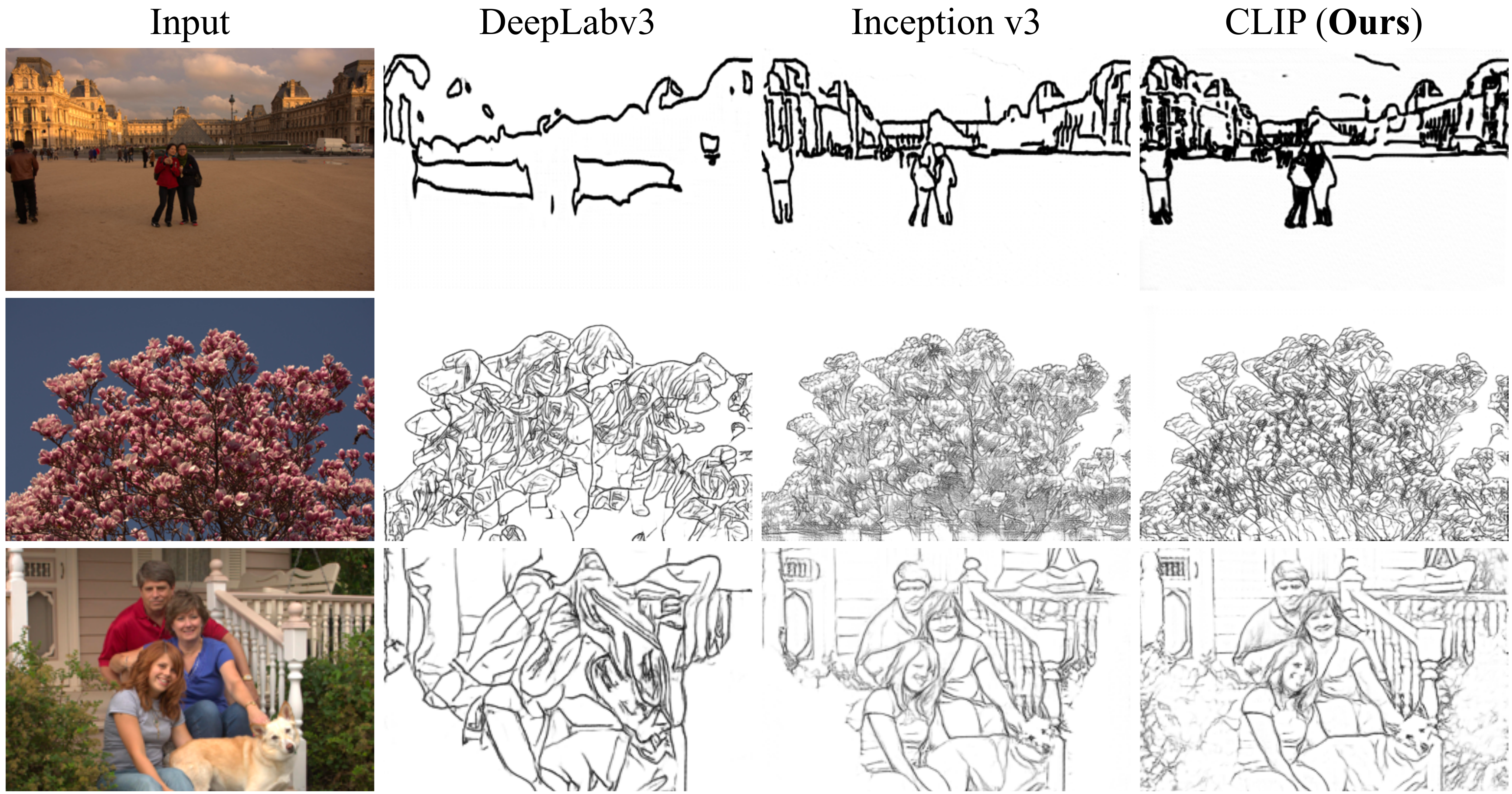}
  \vspace{-0.6cm}
  \caption{Different variants for the semantic loss. From left to right: the input image, output using DeepLabv3 features, output using Inception v3 features, and output using CLIP features (our model).} \label{fig:semanticvariants}
  \vspace{-0.4cm}
\end{figure}

Figure~\ref{fig:semanticvariants} compares semantic features between generated line drawings and input photographs from DeepLabv3~\cite{chen2017deeplab}, Inception v3~\cite{szegedy2016rethinking}, and CLIP ~\cite{clip2021}. Using DeepLabv3 features created confusing drawings. Inception v3 and CLIP features both create reasonable drawings, however we chose the CLIP embedding mainly due to its better handling of lighting, textures, backgrounds, and faces.

\subsection{Network Architectures}

\paragraph{Generators and Discriminators} We use the encoder-decoder generator with $3$ Res-Net blocks~\cite{Simonyan2015} from~\cite{zhu2017unpaired,johnson2016perceptual} for networks $G_A$ and $G_B$. For the disciminators, we use a PatchGAN architecture~\cite{isola2017image} with a receptive field of $70\times70$. The code for these networks are available for academic use under their licenses.

\begin{table}[ht]
  \centering
  \resizebox{0.98\linewidth}{!}{
  \begin{tabular}{ccccccc}
    \toprule
    Layer Type  & Padding & Kernel Size & Stride & Normalization & Activation & Input, Output Channels  \\
\midrule
Conv2D  & $4$ & $7\times7$  & 1 & BatchNorm & ReLU & 768,512  \\
ConvTranspose2D  & $0$ & $4\times4$  & 2 & BatchNorm & ReLU & 512,256  \\
ResNet Block  & $1$ & $3\times3$  & 1 & BatchNorm & ReLU & 256,256  \\
ResNet Block  & $1$ & $3\times3$  & 1 & BatchNorm & ReLU & 256,256  \\
ResNet Block  & $1$ & $3\times3$  & 1 & BatchNorm & ReLU & 256,256  \\
ResNet Block  & $1$ & $3\times3$  & 1 & BatchNorm & ReLU & 256,256  \\
ResNet Block  & $1$ & $3\times3$  & 1 & BatchNorm & ReLU & 256,256  \\
ResNet Block  & $1$ & $3\times3$  & 1 & BatchNorm & ReLU & 256,256  \\
ResNet Block  & $1$ & $3\times3$  & 1 & BatchNorm & ReLU & 256,256  \\
ResNet Block  & $1$ & $3\times3$  & 1 & BatchNorm & ReLU & 256,256  \\
ResNet Block  & $1$ & $3\times3$  & 1 & BatchNorm & ReLU & 256,256  \\
ConvTranspose2D  & $1$ & $3\times3$  & 2 & BatchNorm & ReLU & 256,128  \\
ConvTranspose2D  & $1$ & $3\times3$  & 2 & BatchNorm & ReLU & 128,64  \\
ConvTranspose2D  & $1$ & $3\times3$  & 2 & BatchNorm & ReLU & 64,64  \\
Conv2D  & $3$ & $7\times7$  & 1 & BatchNorm & Tanh & 64,3  \\
\bottomrule
  \end{tabular}
  }
  \caption{Architectures for $G_{Geom}$ which translates ImageNet features into depth maps.} \label{table:ggeom}
  \vspace{-0.2cm}
\end{table}

\paragraph{Depth Networks} To obtain pseudo-ground truth depth maps, we use the output of a pretrained depth prediction system $F$ presented by Miangoleh et al~\cite{Miangoleh2021Boosting} which is built around MiDaS~\cite{Ranftl2019}. These models are available under academic and MIT licenses respectively.

\begin{figure}[t]
  \centering
  \includegraphics[width=\linewidth]{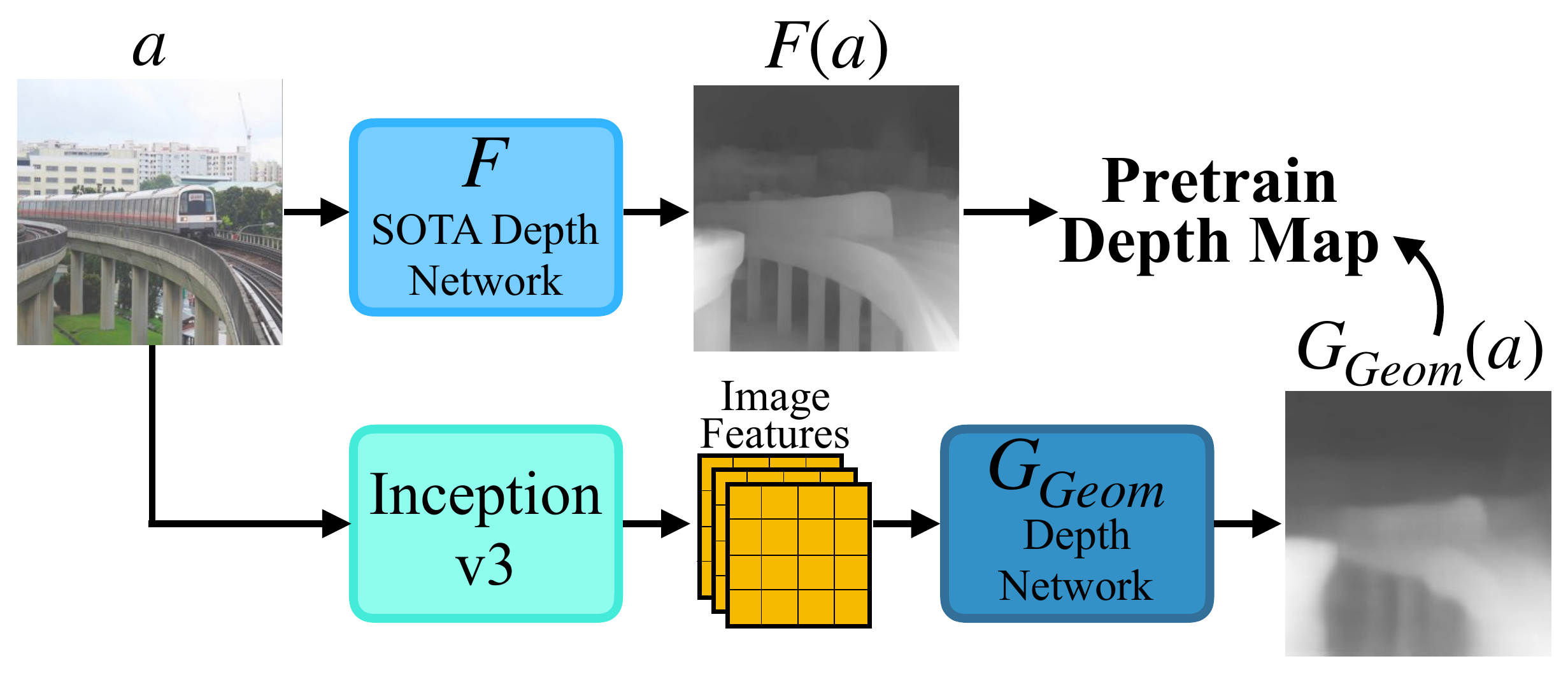}
  \caption{Network $G_{Geom}$ is pretrained to predict depth maps from ImageNet features. $G_{Geom}$ is supervised using state of the art depth prediction network $F$.} \label{fig:pretraindepth} \vspace{-0.3cm}
\end{figure}

The network architecture for image features to depth map network $G_{Geom}$ is based off of the Global Generator presented by Wang et al in pix2pixHD~\cite{wang2018high}. Namely, the beginning layers of the network have been modified to account for the input image features. The individual layers are detailed in Table~\ref{table:ggeom}. $G_{Geom}$ is first pretrained on image features from real photographs to produce depth maps as seen in Figure~\ref{fig:pretraindepth}.

To obtain image features, we use an Inception v3~\cite{szegedy2016rethinking} network which has been pretrained on ImageNet~\cite{deng2009imagenet}. Specifically, we extract features from the Mixed 6b node. We chose this layer for a several reasons. Firstly, previous work has indicated that earlier network features are more important for transfer learning~\cite{kornblith2020s}, and we find this situation applicable to line drawings. Secondly, we conducted a nearest neightbor classification experiment with the ShapeNet dataset~\cite{chang2015shapenet} (terms of use cover research purposes). This dataset consists of many $3$D models of labeled objects. We created renders of these objects, and corresponding line drawings by detecting edges from the depth and normal maps of each render~\cite{saito1990comprehensible}. We then calculated image features for all renders and line drawings at each layer of Inception v3. The nearest neighbor accuracy of the line drawings was then computed with respect to the ImageNet features of the renders. When using features from the middle of the Inception v3~\cite{szegedy2016rethinking} network, the domain gap is reduced. We report nearest neighbor accuracies using features from the penultimate layer and the Mixed 6b node of Inception v3 in Table~\ref{table:acclastlayer}. When using the Mixed 6b node, the nearest neighbor accuracy for line drawings increases to $33.5$\% whereas the nearest neighbor accuracy using features from the penultimate layer is $24.5$\%. In contrast when we perform nearest neighbor classification on rendered images, the accuracy at the penultimate layer is $50.5$\% whereas the accuracy for earlier layers is lower at $38$\%.

\begin{table}[t]
  \centering
  \resizebox{0.9\linewidth}{!}{
  \begin{tabular}{ccc}
    \toprule
      Dataset & 
      Accuracy (penultimate layer) & Accuracy (6b) \\
\midrule
Line Drawings            &     $24.5\%$ & $33.5\%$\\ \hline
Rendered Images          &     $50.5\%$ & $38\%$  \\ \hline
Random                   &     $2.5\%$  & $2.5\%$\\
    \bottomrule
  \end{tabular}
  }
  \caption{Reported nearest neightbor classification accuracies using ImageNet~\cite{deng2009imagenet} features for various methods. The first column uses features from the penultimate layer and the last column uses intermediate features from earlier in the network at the Mixed 6b Node. The last row reports the probability of randomly picking the correct label.} \label{table:acclastlayer} \vspace{-0.6cm}
\end{table}

\paragraph{Semantic Networks}
For the semantic constraint, we use the pretrained CLIP model with a vision transformer (ViT-B/32) base which is presented by OpenAI and is available under an MIT license~\cite{dosovitskiy2020vit, clip2021}.

\subsection{Datasets}

\paragraph{Photograph Datasets}
Our training set for photographs comes from a randomly selected $10,000$ image subset of the Microsoft Common Objects in Context (COCO) dataset~\cite{lin2014microsoft}. The COCO annotations and dataset fall under a Creative Commons 4.0 license, while the images were collected from Flickr and are subject to Flickr terms of use.

We use images from the MIT-Adoke 5k dataset for evaluation. This dataset is available for research purposes under its licenses.

\paragraph{Line Drawing Datasets}

The Contour Drawings dataset consists of $5,000$ images of boundary annotations for various scenes. We use the drawings with width 5. The dataset is available under a Creative Commons 3.0 license.

The Anime Sketch Colorization database consists of $14,224$ pairs of color drawings and sketches of anime characters. In practice, we only use the sketches for the art style. This dataset is available under a Creative Commons 0 (CC0) license. As noted in the main paper, some images may contain sensitive content. For releasing our model, we selected $2256$ training images from this dataset without sensitive content.

OpenSketch consists of $420$ product design sketches of $12$ different objects. Sketches are drawn from multiple viewpoints by different artists. This dataset is available under a Creative Commons 0 (CC0) license.

We also use $207$ artist sketches collected by Cole et al~\cite{cole2008people} to study where people draw lines. We could not find a license for this dataset although it is publicly available and provided by the authors.

\paragraph{Portrait Datasets}
We use two main datasets for evaluating generated portrait drawings. The APDrawings Dataset~\cite{yi2019apdrawinggan} consists of $140$ portrait, line drawing pairs for training (or $420$ pairs after rotational augmentation) and $70$ testing pairs. Although this dataset contains high quality images, it is limited in its size and all portraits are centered with similar poses and lighting. We could not find a license or information on data collection for this dataset, although it is publicly available and linked by the authors.

To create line drawings from portrait photographs under a large variation of poses, lighting, and general differences, we also evaluate our approach using the Helen Facial Feature Dataset~\cite{le2012interactive}. This dataset contains $2000$ training and $330$ test high resolution photos collected for predicting facial annotations from a diverse range of photographs. Images in this dataset are collected from Flickr and subject to their own licenses and copyrights. 

Since we did not have access to the full training set of both portraits and line drawings for style 1, we train our model using the Helen dataset and $88$ line art portraits drawn by Charles Burns~\cite{charlesburns}. Although this second comparison is not exact, we wanted to include a scenario where our method matches the style of drawings created by UPDG.

\begin{table}[t]
  \centering
  \resizebox{0.9\linewidth}{!}{
  \begin{tabular}{ccc}
    \toprule
    Comparison  & Total Judgements & Unique Users  \\
\midrule
Previous Work (Photo)      &     $10,000$  &     $184$ \\ \hline
Previous Work (Portrait)   &     $4,000$  &      $115$ \\ \hline
Ablation                   &     $6,000$    &    $90$ \\ \hline
Geometry Evaluation        &     $6,000$   &     $82$   \\  \hline
Semantics Evaluation       &     $6,000$   &     $135$   \\
    \bottomrule
  \end{tabular}
  }
  \caption{Number of total judgements and unique users for each perceptual study.} \label{table:users} \vspace{-0.5cm}
\end{table}

\subsection{Human subjects and User studies}

Use use Amazon Mechanical Turk for all of our user studies. These studies were conducted with IRB approval. Before participating in the study, users were informed that if they consented, their responses could be presented in meetings and papers, and no personal information would be stored. Users were also told they could decline further participation at any time without adverse consequences.

For most comparisons, we gathered $1000$ total judgements and users viewed up to $100$ images. For comparisons on the APDrawings test dataset, we collected $1200$ judgements and users viewed up to $70$ images (the size of the test dataset). In Table~\ref{table:users} we report the total number of judgements and unique users for each study.

\end{document}